\documentclass{article}
\usepackage{ijcai22}

\usepackage{times}
\usepackage{soul}
\usepackage{url}
\usepackage[hidelinks]{hyperref}
\usepackage[utf8]{inputenc}
\usepackage[small]{caption}
\usepackage{graphicx}
\usepackage{amsmath}
\usepackage{amsthm}
\usepackage{booktabs}
\usepackage{algorithm}
\usepackage{algorithmic}
\usepackage{multibib}
\usepackage{multirow}
\usepackage{tabularx}
\usepackage{soul}
\usepackage{titlesec}
\usepackage{listings}
\usepackage{subcaption}
\usepackage{epstopdf}
\usepackage{xstring}
\usepackage{color}

\title{Handling sign language transcription system with the computer-friendly numerical multilabels}

\author{
Sylwia Majchrowska$^{1,2}$\and
Marta Plantykow$^{3}$\and
Milena Olech$^{3}$\and
\affiliations
$^1$Wroclaw University of Science and Technology, Wrocław, Poland\\
$^2$AI Sweden, Göteborg, Sweden \\
$^3$Intel Technology Poland, Gdańsk, Poland  \\
\emails
sylwia.majchrowska@ai.se,
m.plantykow@gmail.com,
milena.w.olech@gmail.com
}

\begin{document}

\maketitle

\begin{abstract}
    This paper presents our recent developments in the automatic processing of sign language corpora using the Hamburg Sign Language Annotation System (HamNoSys). We designed an automated tool to convert HamNoSys annotations into numerical labels for defined initial features of body and hand positions. Our proposed numerical multilabels greatly simplify annotations' structure without significant loss of gloss meaning. These numerical multilabels can potentially be used to feed the machine learning models, which would accelerate the development of vision-based sign language recognition. In addition, this tool can assist experts in the annotation process and help identify semantic errors. The code and sample annotations are publicly available at \url{https://github.com/hearai/parse-hamnosys}.
\end{abstract}

\section{Introduction}
\label{sec:introduction}

People communicate to exchange information, feelings, and ideas.
Verbal communication is the primary method of human interaction.
Those who speak and listen must use the same type of code (language) to communicate effectively.
Communication outside the Deaf community is a huge challenge, as sign languages (SLs) differ from spoken languages.
The inability to speak makes Deaf people one of the excluded groups in society.

Moreover, different nationalities use different versions of SL, and there is no universal one.
SLs are natural human languages with their own grammatical rules and lexicons.
Therefore, developing an efficient system capable of translating spoken languages into sign languages and vice versa would significantly improve two-way communication.
However, designing and implementing such a machine learning (ML) system requires a large dataset with appropriate annotations.

Traditionally, SL corpora intended for ML training have been annotated at the gloss level.
In the simplest terms, a gloss is a label.
Nevertheless, the same word in different SL can be represented by other gestures.
Therefore, ML methods need large datasets for each SL separately to achieve good results.
Although there is no standard notation system for SL, the Hamburg Sign Language Notation System (HamNoSys) defines a unified set of written symbols that the creators of different SL datasets can use.
This system consists of transcriptions of non--manual and manual features that describe a given sign's shape, orientation, position, and hand movement.

Although HamNoSys is language-agnostic, there are not many ML solutions using this notation system.
First, the grammar of HamNoSys is challenging to understand without specialized linguistic knowledge.
Moreover, using HamNoSys in its raw form in ML-based applications would be extremely difficult. 
An extensive training database is needed to perform this task since using HamNoSys labels would change the classification technique from single-class to multi-class compared to gloss labels.
Furthermore, the inconsistent order of symbols in HamNoSys annotations amongst different datasets makes it almost impossible to process the labels correctly.

To address the problem described above, we created a solution for the automatic processing of SL corpora that transforms HamNoSys labels into numerical labels for carefully defined classes representing the initial body and hand positions.
The numerical labels thus created would support the training of language-agnostic ML models since such multilabel annotations are easier to interpret by users unfamiliar with HamNoSys annotations.
Moreover, the proposed solution allows for further simplification of the annotation structure, making it easier to develop a suitable model.
The presented solution is unique and a publicly available tool that allows arbitrary selection of appropriate classes to describe a specific part of a given gesture.

\section{Related Works}
\label{sec:related_works}

In recent years, work around HamNoSys has focused on creating large multilingual lexicons and simplifying the labeling procedure.
Since the grammar of the notation system is complex and known only to sign language experts, the main effort has been in annotation processing~\cite{Jakub2008InteractiveHN,Justin,skobov-lepage-2020-video}.
Although not all available SL corpora are annotated using HamNoSys, the unified annotation system has attracted the attention of many researchers~\cite{Koller,Dhanjal} because it can significantly simplify multilingual research.

\subsection{HamNoSys-annotated SL Corpora}
\label{Data_Collections}

Deep learning (DL) based approaches are a promising method for the process of sign language translation.
However, they need a large amount of adequately labeled training data to perform well.
As HamNoSys annotations are not language-dependent, they can facilitate cross-language research.
Moreover, merging several multilingual corpora will increase the available data resources.
Below is a comprehensive overview of the existing corpora supplied with HamNoSys labels.

\textbf{GLEX}: The Fachgebärden Lexicon -- Gesundheit and Pflege ~\cite{meinedgs_3} was developed at the German Sign Language Institute of the University of Hamburg between 2004 and 2007 and contains a total of 2330 signs. This dataset is dedicated to technical terms related to health and nursing care.

\textbf{GALEX}: The Fachgebärden Lexicon -- Gärtnerei und Landschaftsbau~\cite{meinedgs_3} was also created at the Hamburg Institute of German Sign Language between 2006 and 2009. This dataset consists of 710 signs related to technical terms for landscaping and gardening.

\textbf{BL}: The Basic Lexicon is part of the multilingual\footnote{British Sign Language, German Sign Language, Greek Sign Language, and French Sign Language} three-year research project DICTA-SIGN~\cite{DictaSign}, carried out since 2009 by a Consortium of European Universities\footnote{Institute for Language and Speech Processing, Universität Hamburg, University of East Anglia, University of Surrey, Laboratoire d’informatique pour la mécanique et les sciences de l’ingénieur, Université Paul Sabatier, National Technical University of Athens, WebSourd}.
About 1k signs are provided for each SL.
The shared list of glosses covers the topic of traveling.

\textbf{CDPSL}: The Corpus-based Dictionary of Polish Sign Language (PJM)~\cite{PJM} was created in 2016 at the Sign Linguistics Laboratory of the University of Warsaw.
The dictionary was developed based on the PJM corpus, which collected the video data of 150 deaf signers who use PJM.
CDPSL allows documenting and describing authentic everyday use of PJM.

\textbf{GSLL}: The Greek Sign Language Lemmas~\cite{GSLL3,GSLL2,GSLL1} is developed by the National Technical University of Athens and supported by the EU research project DICTA-SIGN. The dataset is dedicated to isolated SL recognition and contains 347 glosses signed by two participants, where each sign is repeated between 5 and 17 times.

\subsection{HamNoSys Processing Tools}

In the first decade of the 21st century, Kanis~\textit{et al.}~\cite{Jakub2008InteractiveHN} proposed a HamNoSys editor called an automatic signed speech synthesizer.
This tool can generate a sign animation based on the input HamNoSys label.
The created application can accept the input label in two modes: one designed for direct insertion of HamNoSys symbols and the other with more intuitive graphical interfaces.
The main goal of this solution was to allow an inexperienced annotator to annotate signs correctly.
Unfortunately, it has several limitations, as it contains no more than 300 signs in Czech Sign Language and is based on a domain-specific lexicon.

Power~\textit{et al.}~\cite{Justin} conducted a historical sign language analysis based on 284 multilingual\footnote{American Sign Language, Flemish Sign Language, Mexican Sign Language, French Sign Language, French Belgian Sign Language, Brazilian Sign Language} signs annotated with HamNoSys symbols.
They developed an open-source Python library to facilitate statistical analysis for manipulating HamNoSys annotations.
As a result, they created the first publicly available tool capable of parsing HamNoSys annotations.
However, the proposed parser can only work with sign language data if the HamNoSys labels satisfy strict assumptions. The symbols must appear in the gloss notation in a specific order and be separated by spaces.
Consequently, the tool cannot be used on existing SL corpora without initial preprocessing.

In 2020, an automatic annotation system from video-to-HamNoSys based on the HamNoSys grammar structure was presented~\cite{skobov-lepage-2020-video}.
The proposed solution is based on virtual avatar animations created using the JASigning platform\footnote{https://vh.cmp.uea.ac.uk/index.php/JASigning}.
This approach generated a HamNoSys label based on a given sign animation.
However, the Skobov~\textit{et al.} claimed that the proposed methodology with some modifications could also be used to obtain better results on the real data.

\subsection{ML-based approaches using HamNoSys}

There is a visibly growing interest in sign language recognition from video recordings using ML methods.
Some use HamNoSys labels in a reduced form, as the access to large databases is limited.
The most popular approaches use classification networks.

In 2016, Koller~\textit{et al.}~\cite{Koller} used part of HamNoSys annotations describing the hand orientation modality. They trained classifier on isolated signs in Swiss-German and Danish Sign Language. They applied it to the continuous sign language recognition task on the RWTH-PHOENIX-Weather corpus~\cite{forster-etal-2012-rwth} containing German Sign Language.
This multilingual approach significantly reduced word errors.

Recently, Hidden Markov Models were also used to directly translate multilingual speech into Indian Sign Language~\cite{Dhanjal}.  
In this approach, HamNoSys annotations provided an intermediate step between spoken and sign language. Additional conversion of speech into human-readable text was not needed.
Despite the successful creation of an Indian speech recognition system, it was pointed out that building an optimized model requires a sufficient amount of data in an adequately transcribed format.
This limitation has yet to be overcome.

\section{Hamburg Sign Language Notation System}
\label{sec:hamnosys}

\begin{figure}[!ht]
\centering
  \includegraphics[width=0.5\textwidth]{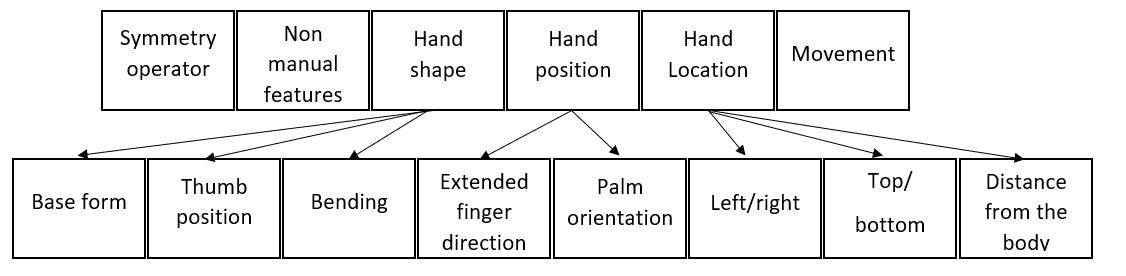}%
  \caption{The structure of HamNoSys consists of six main blocks, four of which are mandatory and two are optional. Additionally, \textbf{Handshape, Hand position} and \textbf{Hand location} blocks are divided into components that specify particular parts of the HamNoSys notation.
  \label{fig.HamNoSysStructure}}
\end{figure}

The Hamburg Sign Language Notation System (HamNoSys) is a \textit{phonetic} transcription system that has been widespread for more than 20 years~\cite{meinedgs_3}.
HamNoSys does not refer to different national finger alphabets and can therefore be used internationally. 
HamNoSys font consists of more than 210 symbols, which encode the initial position of the signer and basic movement of the sign~\cite{Hanke}. 
It can be divided into six basic blocks, as presented in Fig.~\ref{fig.HamNoSysStructure} (upper panel).
The first two out of six blocks -- symmetry operator and non--manual features -- are optional. The remaining four components -- handshape, hand position, hand location, and movement -- are mandatory~\cite{HamNoSysUserGuide}.

Each HamNoSys part mentioned above has the following meaning~\cite{Hanke}:
\begin{itemize}
    \item \textbf{Symmetry operator} (optional): denotes two-handed signs and determines how attributes should be mirrored to the non--dominant hand unless otherwise specified.
    \item \textbf{Non--manual features} (optional): represents non--manual features (e.g., puffed or sucked-in cheeks) that can be used to describe a given sign.
    \item \textbf{Handshape}: refers to the handshape description, which is composed of three subblocks -- \textbf{Base form, Thumb position}, and \textbf{Bending}.
    \item \textbf{Hand position/orientation}: describes the orientation of the hand using two subblocks --
    \textbf{Extended finger direction}, which can specify two degrees of freedom seen from signer’s, birds’ or from-the-right view, and \textbf{Palm orientation}, as third degree of freedom, defined relative to the extended finger direction.
    \item \textbf{Hand location}: is split into three components -- \textbf{Location left/right} specifies x coordinate, \textbf{Location top/bottom} for y coordinate, and \textbf{Distance} (that is skipped if natural) specifies the z coordinate.
    \item \textbf{Movement/Action}: represents a combination of path movements that can be specified as targeted/absolute (location) or relative (direction and size) movements.
\end{itemize}

\section{Decision Tree-based HamNoSys parser}

\begin{figure}[!ht]
\centering
  \includegraphics[width=0.50\textwidth]{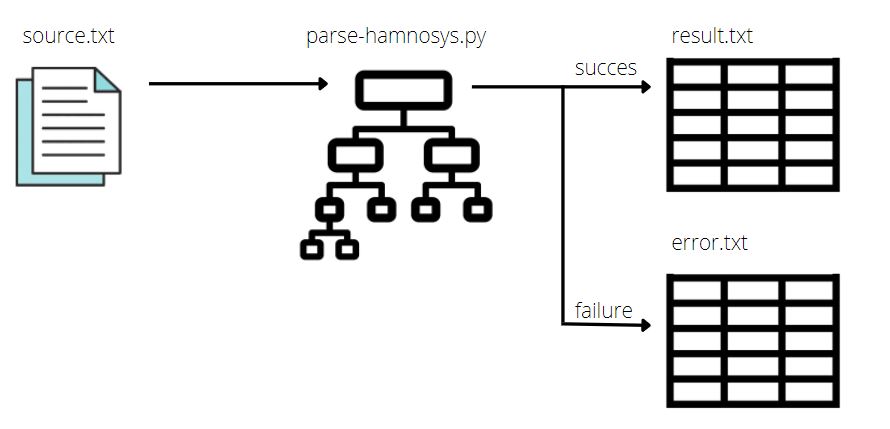}%
  \caption{A schema of action of implemented parser. The source file containing HamNoSys notations is delivered to the parser which is a decision tree and transforms HamNoSys glosses into numeric labels. In case of success, the data are written to the result text file, otherwise error file is updated.}
  \label{fig.schemat}
\end{figure}

The main goal of the parser is to translate a label representing a SL gloss, written in HamNoSys format, into a form that can be used for DL-based classification.
Since the structure of the HamNoSys grammar can be described as a decision tree~\cite{skobov-lepage-2020-video}, we used the method to decompose the notation into numerical multilabels for the defined classes.
The parser logic implements the rules underlying data with the sequential structure to analyze a series of symbols.
It matches each symbol with the class that describes it (while assigning it the appropriate number) or removes it.
The fig.~\ref{fig.schemat} shows a diagram of how a parser works.

\subsection{Methodology}

As described in Section~\ref{sec:hamnosys}, HamNoSys labels can be represented by blocks. 
In our implementation, four blocks (symmetry operator, location left/right, location top/bottom, distance from the body) refer to the overall human posture
In comparison, five blocks (handshape base form, handshape thumb position, handshape bending, hand position extended finger direction, and hand position palm orientation) relate to a single hand.

\begin{figure}[!ht]
\centering
  \includegraphics[width=0.54\textwidth]{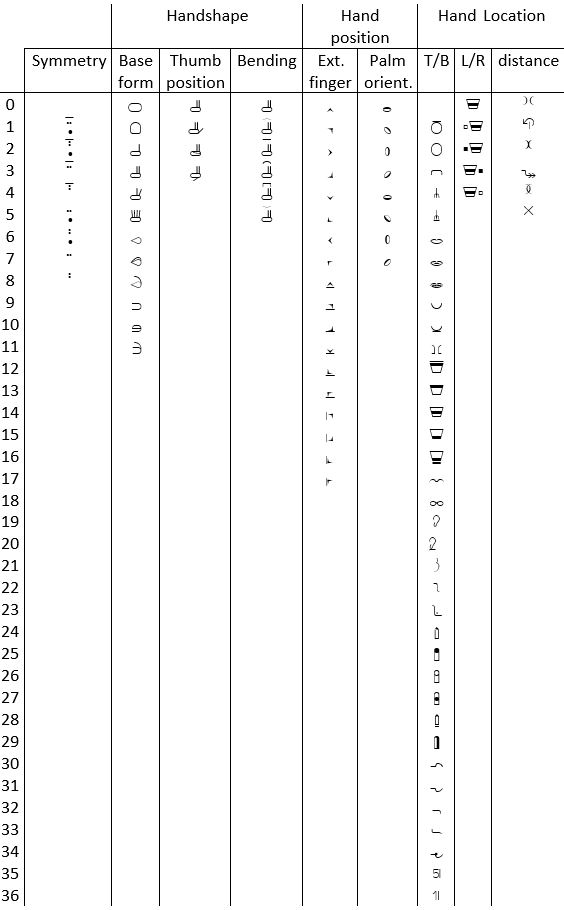}%
  \caption{Parser classes numerical values and its counterpart character in HamNoSys font. The thumb position and bending are presented in combination with the handshape base form symbol to increase its readability.}
  \label{fig.ClassesAll}
\end{figure}

Furthermore, both hands (dominant and non--dominant) can be involved in each sign.
Up to two HamNoSys symbols describing a single class can be assigned for each hand.
To properly store all symbols, the classes related to a single hand are repeated four times as the primary and secondary descriptions for the dominant hand and the primary and secondary descriptions for the non--dominant hand.
Moreover, we added one extra class that indicates if the sign description starts from a relaxed hand sign.
As a result, together with classes describing overall human posture, the parser considers 25 classes when analyzing HamNoSys labels.
The fig.~\ref{fig.ClassesAll} presents the numerical values and assigns HamNoSys symbols to them.

Class \textbf{Symmetry operator}, describes how the description of the dominant hands maps to the non--dominant hand using nine numbers.
A value of class \textbf{NonDom first} equals 1 if the sign description starts from a relaxed hand sign, otherwise is set to 0.

The following 20 classes contain primary and secondary descriptions of five features assigned separately for dominant and non--dominant hands.
Secondary classes will be assigned if the $\backslash$ operator is used. For example, if A$\backslash$B construct is used, symbol A will be assigned to the primary class, and symbol B will be assigned to the secondary class.

The \textbf{Handshape base form} class describes the base form of a handshape in the initial posture.
This class value can be in the range of 0 to 11.
The appearance of the class symbol is expected right after the symmetry operator.
Only three symbols: relaxed hand, \~{} or [ can occur in between.
The first one, indicating a relaxed hand, will be assigned a numeric label, and the other two will be omitted.
An error will occur if this symbol is not found in the expected place.

The \textbf{Handshape thumb position} class describes the position of a thumb in the initial posture and has four possibilities.
This class symbol can only be used right after the \textit{handshape base form} symbol or \textit{bending} symbol.
Otherwise, an error will occur.
The \textbf{Handshape bending} class describes the hand bending in the initial posture.
This class symbol is in the range 0 to 5 and can be used only right after the \textit{handshape base form} symbol or \textit{handshape thumb position} symbol.
Otherwise, an error will occur.
Fig.~\ref{fig.ClassesAll} presents the thumb position and bending combined with the handshape base form icon to increase its readability.

The \textbf{Hand position extended finger direction} class specifies two degrees of freedom.
This class value can be in the range of 0 to 17.
This class must occur at least once in the HamNoSys label.
Otherwise, an error will occur since this class is mandatory.
The \textbf{Hand position palm orientation} class is in the range 0 to 7 and specifies the third degree of freedom.
Like the previous one, this class is mandatory.
The symbol must be found in the HamNoSys label at least once - if not, an error will occur.

Three remaining classes, \textbf{Hand location L/R} (Left/Right) \textbf{Hand location T/B} (Top/Bottom) and \textbf{Hand location distance}, describe the hand location, using x, y and z coordinates respectively.
The first is 0 to 4, the second uses 37 symbols, and the last class values are between 0 and 5.
For \textbf{Hand location L/R} class, the symbol position is analyzed in relation to the \textbf{Hand location T/B} symbol (see Fig.~\ref{fig.ClassesAll}).

As previously described, there are three mandatory symbol classes the parser shall find in the HamNoSys label: \textit{hand position base form}, \textit{hand extended finger direction}, and \textit{hand palm orientation}.
If any of those classes are not found, an error will occur.

As suggested in sources~\cite{Hanke}, if the location information is missing, the default \textit{neutral} values are assigned automatically, meaning 0 for distance, 0 for left/right, and 14 for top/bottom.
All classes not found in the HamNoSys label are marked as \textit{NaN} to distinguish them easily.
An error when parsing a particular class will be indicated by a negative number assigned to the class.

\subsection{Example usage}

The open-source code is implemented in Python programming language, making it more flexible and user-friendly and allowing it to incorporate commonly used Python libraries such as Pandas.

The package consists of two main scripts: \textit{parse-hamnosys}, which is the primary driver, and \textit{hamnosys\_dicts}, which is the dictionary file.
The \textit{parse-hamnosys} script can be used to convert a HamNoSys encoding into numerical labels.
It requires arguments indicating the source file containing HamNoSys notation and destination files that separately contain successfully and not successfully parsed results.
Moreover, two optional arguments specify the names of columns in the input and output text files.

Listing~\ref{lst:parse} gives an example of the command to use the parser with the mentioned above parameters.
As a result, two files, \textit{source.txt} and \textit{error.txt}, will be generated.
The first of them contains correctly parsed glosses with their classes. The other one includes glosses that were not parsed correctly. 

\begin{lstlisting}[language=bash, caption={Basic example of calling \textit{parse-hamnosys} script.},label={lst:parse}]
python3 parse-hamnosys.py \
    --src_file source.txt \
    --dst_file result.txt \
    --err_file error.txt
\end{lstlisting}

Table~\ref{tab:results} presents a few examples from two different datasets.
The example 1 stands for word $E \Pi I \Sigma K E \Upsilon A Z \Omega$ (encoded as \textit{7 0 0 0 0 0 5 2 0 4 NaN 0 0 0 1 6 2 0 4 NaN NaN 0 13 0}, which each position of numerical label is related to specific HamNoSys block) 
and comes from GSLL dataset. The second example comes from Basic Lexicon dataset and stands for word \textit{know}.
The term NaN emphasizes that a given feature of a gloss was not specified in the HamNoSys string.

\begin{table}[!h]
\begin{tabular}{|lll|l|l|}
\hline
\multicolumn{3}{|l|}{Class} & Ex. 1 & Ex. 2 \\ \hline
\multicolumn{1}{|l|}{\multirow{2}{*}{}} & \multicolumn{2}{l|}{Symmetry} & 7 & 0 \\ \hline
\multicolumn{1}{|l|}{\multirow{2}{*}{}} & \multicolumn{2}{l|}{NonDom first} & 0 & 0 \\ \hline
\multicolumn{1}{|l|}{\multirow{5}{*}{Dom. 1}} & \multicolumn{1}{l|}{\multirow{3}{*}{Shape}} & Base form & 0 & 0 \\ \cline{3-5} 
\multicolumn{1}{|l|}{} & \multicolumn{1}{l|}{} & Thumb & 0 & 1 \\ \cline{3-5} 
\multicolumn{1}{|l|}{} & \multicolumn{1}{l|}{} & Bending & 0 & 0 \\ \cline{2-5} 
\multicolumn{1}{|l|}{} & \multicolumn{1}{l|}{\multirow{2}{*}{Position}} & Ext.finger dir. & 0 & 0 \\ \cline{3-5} 
\multicolumn{1}{|l|}{} & \multicolumn{1}{l|}{} & Palm & 5 & 4 \\ \hline
\multicolumn{1}{|l|}{\multirow{5}{*}{Dom. 2}} & \multicolumn{1}{l|}{\multirow{3}{*}{Shape}} & Base form & 2 & NaN \\ \cline{3-5} 
\multicolumn{1}{|l|}{} & \multicolumn{1}{l|}{} & Thumb & 0 & NaN \\ \cline{3-5} 
\multicolumn{1}{|l|}{} & \multicolumn{1}{l|}{} & Bending & 4 & NaN \\ \cline{2-5} 
\multicolumn{1}{|l|}{} & \multicolumn{1}{l|}{\multirow{2}{*}{Position}} & Ext.finger dir. & NaN & 7 \\ \cline{3-5} 
\multicolumn{1}{|l|}{} & \multicolumn{1}{l|}{} & Palm & NaN & 2 \\ \hline
\multicolumn{1}{|l|}{\multirow{5}{*}{Ndom. 1}} & \multicolumn{1}{l|}{\multirow{3}{*}{Shape}} & Base form & 0 & NaN \\ \cline{3-5} 
\multicolumn{1}{|l|}{} & \multicolumn{1}{l|}{} & Thumb & 0 & NaN \\ \cline{3-5} 
\multicolumn{1}{|l|}{} & \multicolumn{1}{l|}{} & Bending & 0 & NaN \\ \cline{2-5} 
\multicolumn{1}{|l|}{} & \multicolumn{1}{l|}{\multirow{2}{*}{Position}} & Ext.finger dir. & 1 & NaN \\ \cline{3-5} 
\multicolumn{1}{|l|}{} & \multicolumn{1}{l|}{} & Palm & 6 & NaN \\ \hline
\multicolumn{1}{|l|}{\multirow{5}{*}{Ndom 2}} & \multicolumn{1}{l|}{\multirow{3}{*}{Shape}} & Base form & 2 & NaN \\ \cline{3-5} 
\multicolumn{1}{|l|}{} & \multicolumn{1}{l|}{} & Thumb & 0 & NaN \\ \cline{3-5} 
\multicolumn{1}{|l|}{} & \multicolumn{1}{l|}{} & Bending & 4 & NaN \\ \cline{2-5} 
\multicolumn{1}{|l|}{} & \multicolumn{1}{l|}{\multirow{2}{*}{Position}} & Ext.finger dir. & NaN & NaN \\ \cline{3-5} 
\multicolumn{1}{|l|}{} & \multicolumn{1}{l|}{} & Palm & NaN & NaN \\ \hline
\multicolumn{1}{|l|}{} & \multicolumn{1}{l|}{\multirow{3}{*}{Location}} & x & 0 & 4 \\ \cline{1-1} \cline{3-5} 
\multicolumn{1}{|l|}{} & \multicolumn{1}{l|}{} & y & 13 & 3 \\ \cline{1-1} \cline{3-5} 
\multicolumn{1}{|l|}{} & \multicolumn{1}{l|}{} & z & 0 & 0 \\ \hline
\end{tabular}
\caption{The example of the final numerical labels created by the parser for two glosses (separate words) -- $E \Pi I \Sigma K E \Upsilon A Z \Omega$ (Ex.1) and \textit{know} (Ex.2) -- from two databases. The NaN occurs when no symbol in the HamNoSys source string encodes the manual or non--manual feature (i.e., the Ex.2 is described only for the dominant hand).}
\label{tab:results}
\end{table}

\subsection{Error handling}
Decision tree-based parser implementation predefines possible HamNoSys label formats. 
If the parser does not recognize the order of symbols in the HamNoSys label, an error will be announced by filling the given column with a negative value.

We used the open part of each database mentioned in Section~\ref{Data_Collections} to analyze the parser effectiveness.
The gathered collection of datasets consists of around ten hours of videos accompanied by 11831 glosses, where 7095 are unique.

Table~\ref{errors} presents the number of successfully parsed entries for each dataset analyzed.
The \textit{\# glosses} column represents the number of all glosses passed to the parser as an input.
The \textit{\# correct} column contains the number of correctly parsed glosses.
In contrast, the \textit{\# errors} column shows the number of HamNoSys labels in which the order of symbols was not properly recognized.
The parser effectiveness $\mathrm{\mathbf{\eta_p}}$ was calculated as the percentage of entries that were correctly decomposed and parsed by the parser. 

\begin{table}[ht]
\centering
\caption{Number of successfully parsed entries for each dataset.}
\begin{tabular}{c c c c c}
\hline
\textbf{Dataset} & \textbf{\#} & \textbf{\#} & \textbf{\#}    & $\mathrm{\mathbf{\eta_p}}$ \\
 \textbf{name} & \textbf{glosses} & \textbf{parsed} & \textbf{errors}    & \textbf{(\%)} \\
\hline
GALEX               & 568       & 561       & 7         &98.77   \\
GLEX                & 829       & 778       & 51        &93.85   \\
CDPSL               & 2835      & 2828      & 7         &99.75   \\
BL                  & 4123      & 3907      & 216       &94.76   \\
GSLL                & 3476      & 3316      & 160       &95.40   \\
\hline
\end{tabular}
\label{errors}
\end{table}

For each of the analyzed datasets parser successfully decomposed over 93\% of entries, reaching the lowest effectiveness when analyzing the GLEX dataset that consists of 829 entries (93.85\%) and the best result for CDPSL (99.75\%) analyzing 2835 notations.

As previously mentioned, the parser's main logic is based on a concept of a predefined decision tree, so the possible order of symbols in HamNoSyS notation is predefined. Additional manual analysis of notations that were not correctly processed by the parser demonstrated that in most cases, a label was not processed correctly as the predefined order of symbols was not applied. 
In some cases, such as duplicated hand shape bending symbol, additional logic was implemented to handle exceptions (remove excessive symbol).
Unfortunately, it was impossible to forecast and serve all possible deviations. 

The highest processing efficiency was reached for the CDPSL dataset, which can be connected with the time that the authors of the parser have spent analyzing this particular dataset and all possible workarounds that will help preserve the predefined order of symbols.
The difference in obtained results proves that the expected order of symbols is not equally preserved in all of the datasets, meaning that all of the datasets were annotated in a different manner.
If an error occurs, it is highly advised to manually analyze the order of symbols in the HamNoSys string placed in \textit{error.txt} file.





\section{Data reduction influence analysis}
\label{Usefulness}

As a final analysis, we performed backward decoding from the assigned numeric multilabels to the correct glosses.
In this study, we evaluated the impact of information reduction caused by the proposed HamNoSys label encoding methodology.

It was assumed that the decoding process was successful if the parser assigned one HamNoSys label to reach the tested gloss.
If it assigned more than one label, the verification was considered unsuccessful.
Due to database diversity and significant data differences, the results were verified on each of them separately.
The lowest decoding efficiency $\mathrm{\mathbf{\eta_d}}$ was observed for Basic Lexicon, which is a set that consists of four SLs.
For this database, the number of glosses assigned to a single HamNoSys label (and thus misidentified) reached the highest value of 12. 
It is the same as in the context of spoken language, where there are similar words, 
there may be similar gestures that describe different glosses, 
especially in multilingual databases such as BL. 
Moreover, the primary purpose of creating BL was to identify common signs between the four sign languages.

Table~\ref{tab.resultse} summarizes the influence of data reduction on each dataset. The \textit{\# unique glosses} column indicates the total number of distinctive glosses processed by the parser and assigned to HamNoSys labels. The \textit{\# singly labeled} and \textit{\# repeated} columns show the number of singly and multiply decoded glosses, respectively. Finally, the calculated high decoding efficiency $\mathrm{\mathbf{\eta_d}} > 83\%$ proves that the parser can correctly recognize an individual gloss from different SLs. 
It is worth noting that some level of similarity (when the signatures of different glosses are similar) is inevitable due to the homonymy of language. 
We conclude that data reduction has no significant negative impact on possible invert decoding from numerical multilabels into glosses. 

\begin{table}[!h]
\centering
\caption{A summary of the impact of data reduction on individual datasets that presents a total number of unique glosses in each dataset, number of glosses singly assigned to a unique HamNoSys label (consequently gloss), and the number of repeated assignments. The last column shows the parser's capability to recognize a single isolated gloss.}
\begin{tabular}{c c c c c}
\hline
\textbf{Dataset} & \textbf{\# unique} & \textbf{\# singly} & \textbf{\#}    & $\mathrm{\mathbf{\eta_d}}$ \\
 \textbf{name} & \textbf{glosses} & \textbf{labelled} & \textbf{repeated}    & \textbf{(\%)} \\
\hline
GALEX               & 514       & 484       & 30        &94.16   \\
GLEX                & 723       & 684       & 39        &94.61   \\
CDPSL               & 2480      & 2259      & 221       &91.09   \\
BL                  & 3078      & 2580      & 498       &83.82   \\
GSLL                & 300       & 283       & 17        &94.33   \\
\hline
\end{tabular}
\label{tab.resultse}
\end{table}

\section{Conclusion}

Nowadays, several lexical corpus collections include HamNoSys annotations.
However, the way signs are annotated is not standardized.
We decided to implement the HamNoSys parser as universally as possible to leverage and combine existing annotation efforts from different corpora. 
The parser focuses on converting HamNoSys symbols into a spatial-positional representation encoded by numerical multilabels to make existing corpora in this format useful for ML-based tasks and more straightforward to understand for non--linguistic researchers.

The proposed parser reduces the amount of data stored in the original HamNoSys character since it omits some data, such as movement (analyzing only the initial gloss position) or finger-related details.
Nevertheless, the essential characteristic of the sign is preserved.
We have also proved that backward decoding of the gloss is possible with efficiency above 83\%.

The main benefit of this solution is that the created numerical labels can be used as input for multi-headed classification networks of the signer's initial position and then for recognition of targeted words in spoken language. 
This way of representing HamNoSys also simplifies its structure even further, e.g., encoding only the shape, orientation, and position of the dominant hand. 
We believe that the developed tool will contribute to future research efforts to create a fully functional sign language-agnostic translator.

\section*{Acknowledgements}

The five-month non--profit educational project HearAI was organized by
Natalia Czerep, Sylwia Majchrowska, Agnieszka Mikołajczyk, Milena Olech, Marta Plantykow, Żaneta Lucka Tomczyk.
The main goals were to work on sign language recognition using HamNoSys and increase public awareness of the Deaf community.
The authors acknowledge other team members: Natalia Czerep, Maria Ferlin, Wiktor Filipiuk, Grzegorz Goryl, Agnieszka Kamińska, Alicja Krzemińska, Adrian Lachowicz, Agnieszka Mikołajczyk, Patryk Radzki, Krzysztof Bork-Ceszlak, Alicja Kwasniewska, Karol Majek, Jakub Nalepa, Marek Sowa
who contributed to the project.
The authors acknowledge the infrastructure and support of Voicelab.ai in Gdańsk and the financial support of the Polish Development Fund (PFR) Foundation.

\bibliographystyle{named}
\bibliography{ijcai22}

\end{document}